\documentclass{bmvc2k}

\usepackage{enumitem}
\usepackage{amsfonts}
\usepackage{algorithm}
\usepackage[noend]{algpseudocode}
\usepackage[normalem]{ulem}
\usepackage{changepage}
\usepackage{tabulary}
\usepackage{tabularx}
\usepackage{multirow}
\usepackage{hhline}
\usepackage[title]{appendix}

\makeatletter
\def\BState{\State\hskip-\ALG@thistlm}
\makeatother

\definecolor{orange}{RGB}{255, 165, 0}

\title{RISE: Randomized Input Sampling for Explanation of Black-box Models}

\addauthor{Vitali Petsiuk}{vpetsiuk@bu.edu}{1}
\addauthor{Abir Das}{dasabir@bu.edu}{1}
\addauthor{Kate Saenko}{saenko@bu.edu}{1}

\addinstitution{
 Boston University\\
 Boston, USA
}

\runninghead{Petsiuk, Das, Saenko}{RISE: Randomized Input Sampling for Explanation}

\def\eg{\emph{e.g}\bmvaOneDot}

\def\etal{\emph{et al}\bmvaOneDot}
\def\ie{\emph{i.e.}}
\def\etc{\emph{etc}\bmvaOneDot}

\begin{document}
\maketitle
    
\begin{abstract}
    Deep neural networks are being used increasingly to automate data analysis and decision making, yet their decision-making process is largely unclear and is difficult to explain to the end users. In this paper, we address the problem of Explainable AI for deep neural networks that take images as input and output a class probability. We propose an approach called RISE that generates an importance map indicating how salient each pixel is for the model's prediction. In contrast to white-box approaches that estimate pixel importance using gradients or other internal network state, RISE works on black-box models. It estimates importance empirically by probing the model with randomly masked versions of the input image and obtaining the corresponding outputs. We compare our approach to state-of-the-art importance extraction methods using both an automatic deletion/insertion metric and a pointing metric based on human-annotated object segments. Extensive experiments on several benchmark datasets show that our approach matches or exceeds the performance of other methods, including white-box approaches.
\end{abstract}
    
\section{Introduction}
\label{sec:intro}
\vspace{-1mm}

Recent success of deep neural networks has led to a remarkable growth in Artificial Intelligence (AI) research.
In spite of the success, it remains largely unclear how a particular neural network comes to a decision, how certain it is about the decision, if and when it can be trusted, or when it has to be corrected.
In domains where a decision can have serious consequences (\eg, medical diagnosis, autonomous driving, criminal justice \etc), it is especially important that the decision-making models are transparent.
There is extensive evidence for the importance of explanation towards understanding and building trust in cognitive psychology~\cite{Lombrozo2006Structure}, philosophy~\cite{Lombrozo2011Instrumental} and machine learning~\cite{Dzindolet2003Role, Ribeiro2016Should, Lipton2016Mythos} research. 
In this paper, we address the problem of \textit{Explainable AI}, \ie, providing explanations for the artificially intelligent model's decision. Specifically, we are interested in explaining classification decisions made by deep neural networks on natural images.

\begin{figure}[!t]
	\vspace{-2mm}
    \centering
    \subfigure[\small{Sheep - $26\%$, Cow - $17\%$}]{
    	\label{fig:Original_Sheep}
    	\includegraphics[width=0.30\linewidth]
    {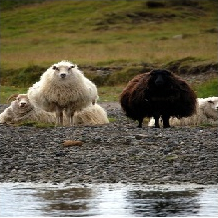}}
    \subfigure[\small{Importance map of `\textit{sheep}'}]{
    	\label{fig:Sheep_Importance}
        \includegraphics[width=0.30\linewidth]
        {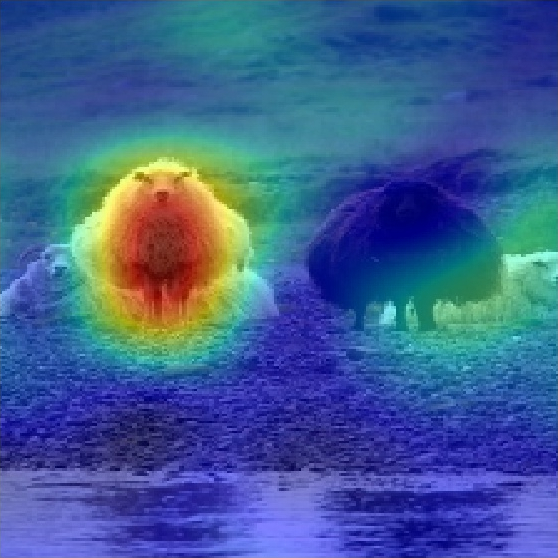}}
    \subfigure[\small{Importance map of `\textit{cow}'}]{
        \label{fig:Cow_Importance}
    	\includegraphics[width=0.30\linewidth]
    	{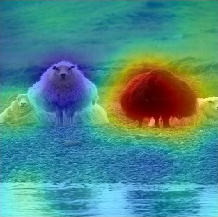}}\\
    \subfigure[\small{Bird - $100\%$, Person - $39\%$}]{
        \label{fig:Original_Bird}
        \includegraphics[width=0.30\linewidth]
        {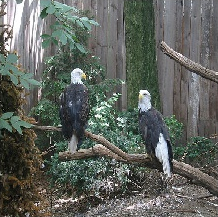}}
    \subfigure[\small{Importance map of `\textit{bird}'}]{
        \label{fig:Bird_Importance}
        \includegraphics[width=0.30\linewidth]
        {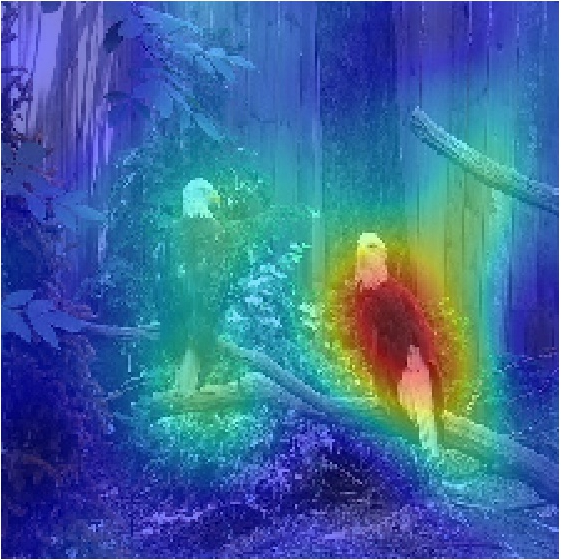}}
    \subfigure[\small{Importance map of `\textit{person}'}]{
        \label{fig:Person_Importance}
        \includegraphics[width=0.30\linewidth]
        {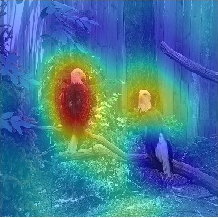}}\\
    \caption{\small Our proposed RISE approach can explain why a black-box model (here, ResNet50) makes classification decisions  by generating a pixel importance map  for each decision (redder is more important). For the top image, it reveals that the model only recognizes the white sheep and confuses the black one with a cow; for the bottom image it confuses parts of birds with a person. (Images taken from the PASCAL VOC dataset.)}
    \vspace{-3mm}
\end{figure}

Consider the prediction of a popular image classification model (ResNet50 obtained from~\cite{zhang2017EB}) on the image depicting several sheep shown in Fig.~\ref{fig:Original_Sheep}. 
We might wonder, why is the model predicting the presence of a cow in this photo? Does it see all sheep as equally sheep-like?
An explainable AI approach can  provide answers to these questions, which in turn can help fix such mistakes. In this paper, we take a popular approach of generating a \textit{saliency} or \textit{importance} map that shows how important each image pixel is for the network's prediction.
In this case, our approach reveals that the ResNet model confuses the black sheep for a cow (Fig.~\ref{fig:Cow_Importance}), potentially due to the scarcity of black colored sheep in its training data.
A similar observation is made for the photo of two birds (Fig~\ref{fig:Original_Bird}) where the same ResNet model predicts the presence of a bird and a person.
Our generated explanation reveals that the left bird provides most of the visual evidence for the `person' class.

\begin{figure}[!t]
    \vspace{-2mm}
    \centering
    \subfigure[Input]{
        \label{fig:Original}
        \includegraphics[width=0.23\linewidth]{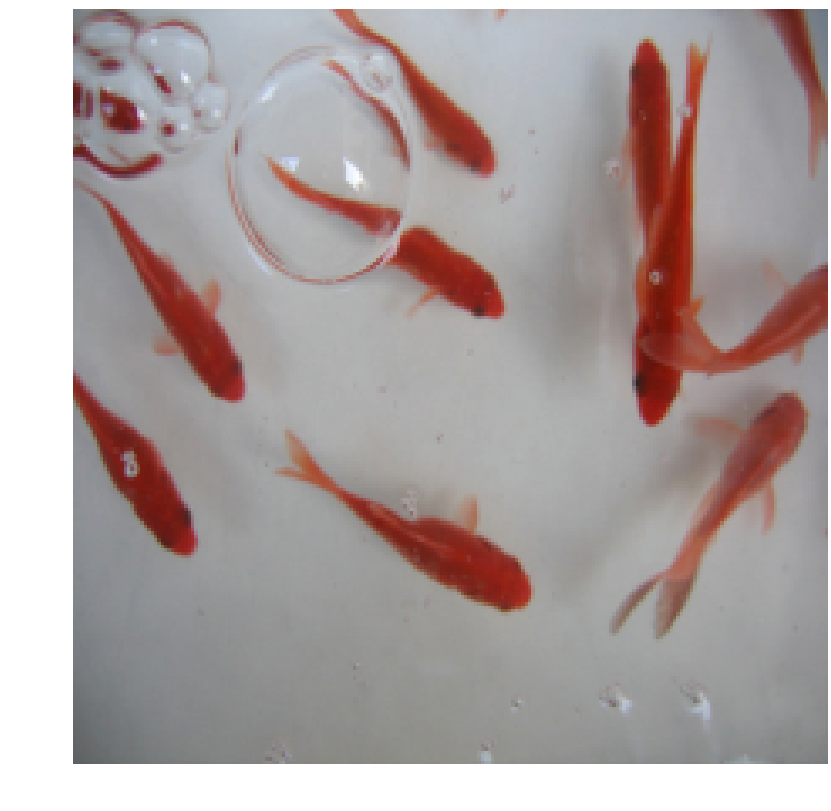}}
    \subfigure[RISE (ours)]{
    \label{fig:Explanation}
    \includegraphics[width=0.23\linewidth]{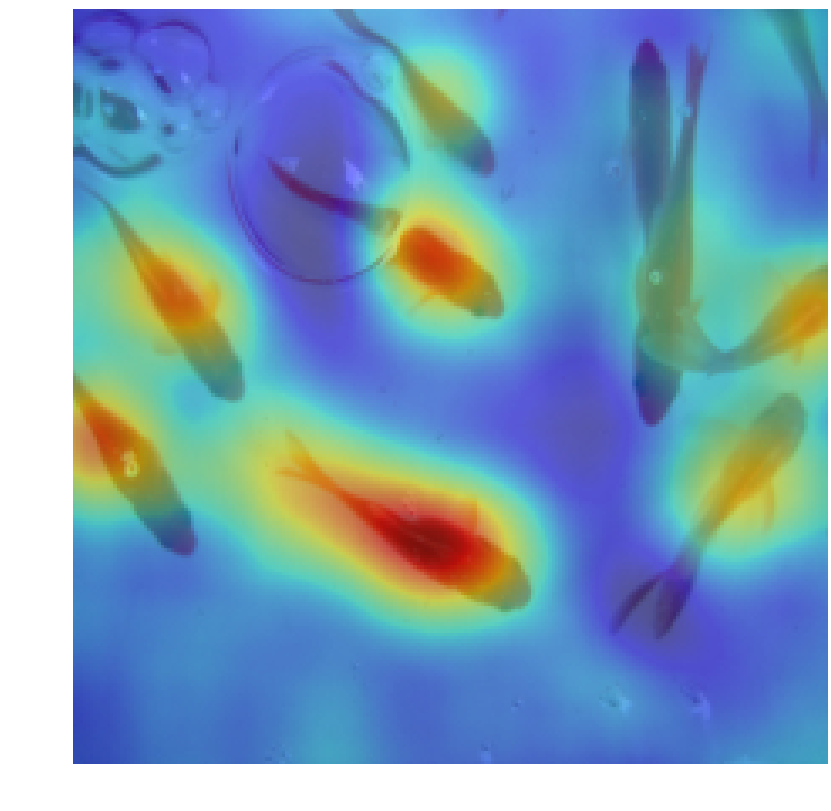}}
    \subfigure[GradCAM]{
        \label{fig:Gradcam}
        \includegraphics[width=0.23\linewidth]{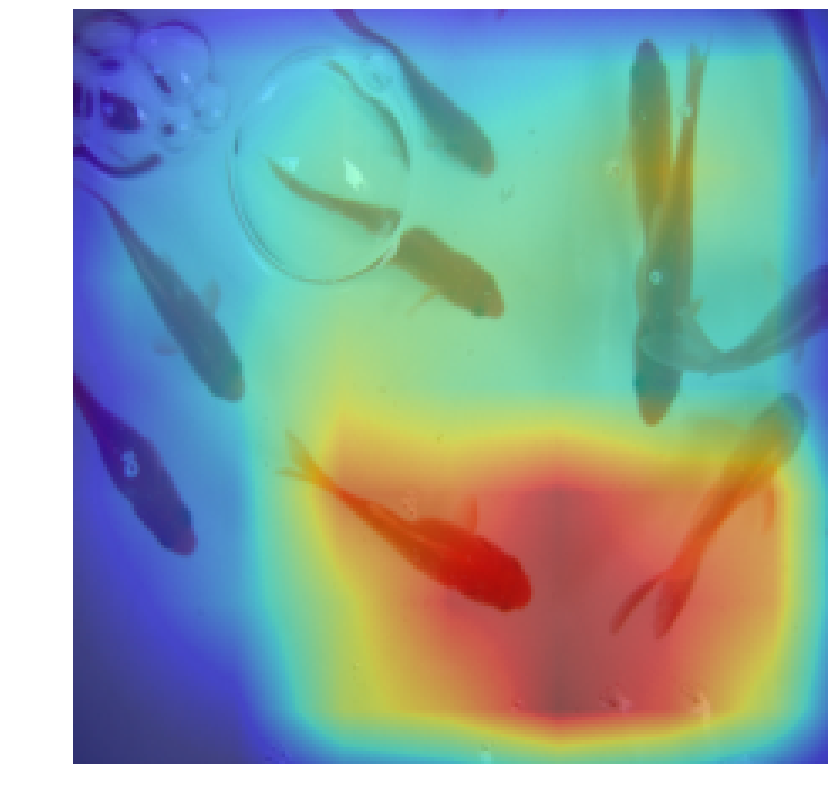}}
    \subfigure[LIME]{
        \label{fig:LIME}
        \includegraphics[width=0.23\linewidth]{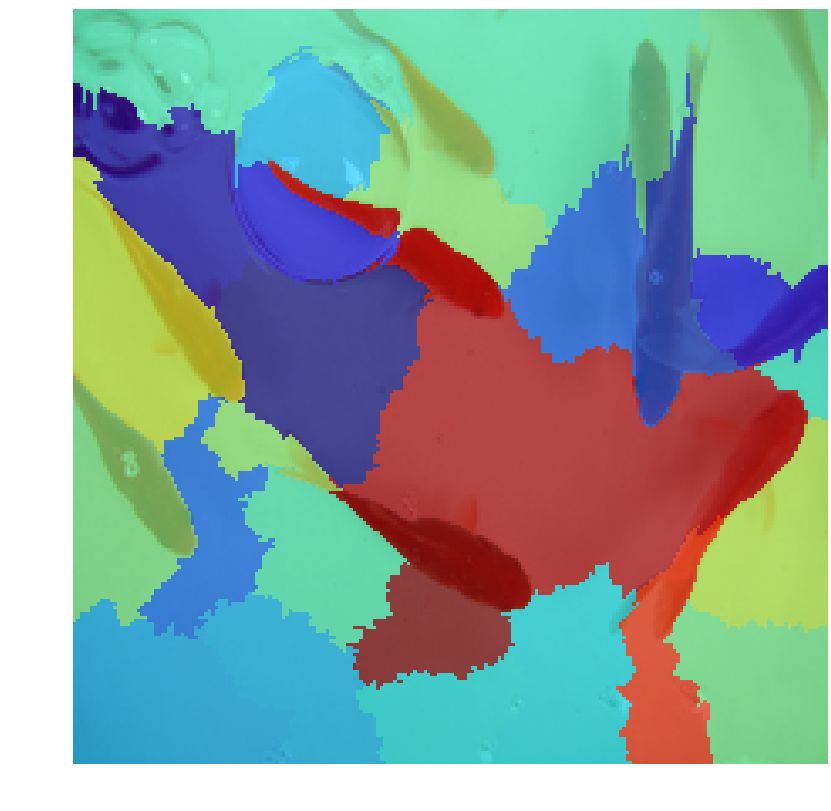}}\\
    \subfigure[Image during deletion]{
        \label{fig:RISEmask}
        \includegraphics[width=0.23\linewidth]{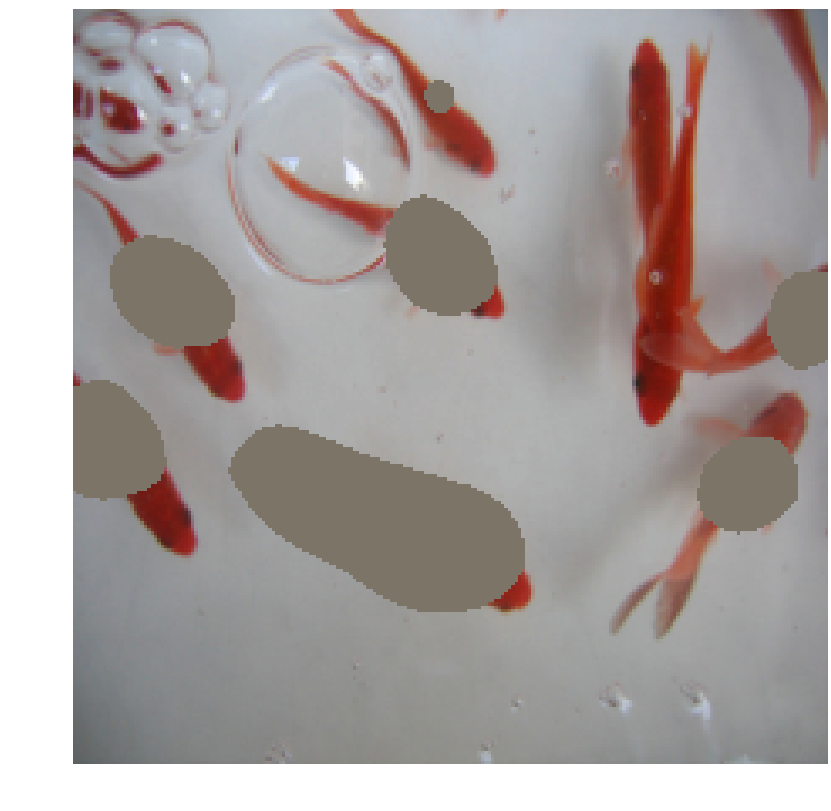}}
    \subfigure[RISE-Deletion]{
        \label{fig:SelExplanation}
        \includegraphics[width=0.23\linewidth]{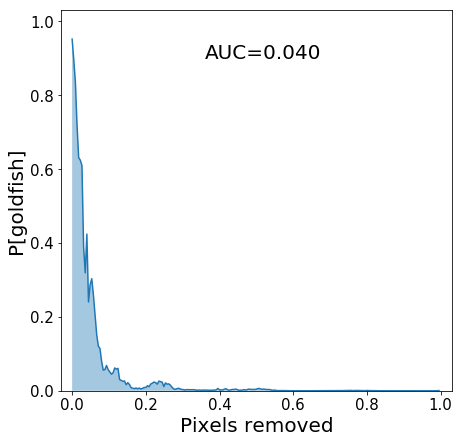}}
    \subfigure[GradCAM-Deletion]{
        \label{fig:SelFinetuned}
        \includegraphics[width=0.23\linewidth]{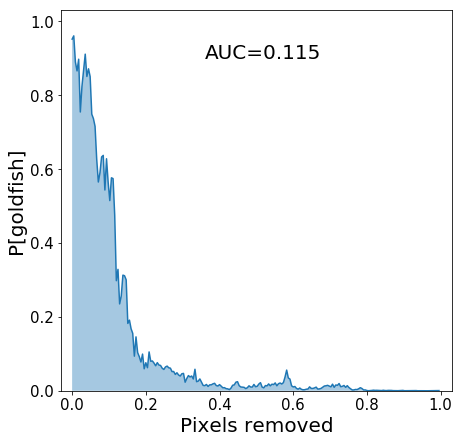}}
    \subfigure[LIME-Deletion]{
        \label{fig:SelGradCam}
        \includegraphics[width=0.23\linewidth]{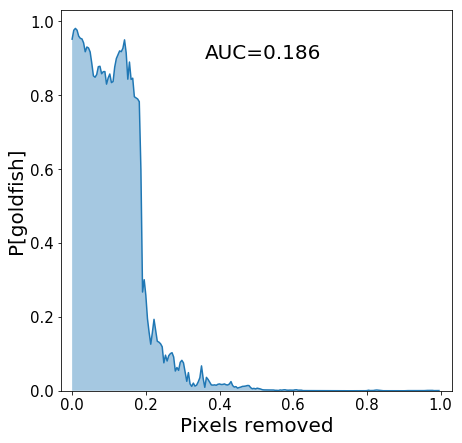}}
    \caption{\small Estimation of importance of each pixel by RISE and other state-of-the-art methods for a base model's prediction along with `deletion' scores (AUC).
    The top row shows an input image (from ImageNet) and saliency maps produced by RISE, Grad-CAM~\cite{Selvaraju2017Gradcam} and LIME~\cite{Ribeiro2016Should} with ResNet50 as the base network (redder values indicate higher importance).
    The bottom row illustrates the deletion metric: salient pixels are gradually masked from the image (\ref{fig:RISEmask}) in order of decreasing importance, and the probability of the `goldfish' class predicted by the network is plotted \textit{vs.} the fraction of removed pixels. In this example, RISE provides more accurate saliency and achieves the lowest AUC.}
    \label{fig:ExplanationAndSelectivity}
    \vspace{-3mm}
\end{figure}

Existing methods~\cite{Simonyan2013Deep, Yosinski2014Understanding, Nguyen2016Synthesizing, Zhou2016Learning, Selvaraju2017Gradcam, zhang2017EB, Fong2017Interpretable} compute importance for a given base model (the one being explained) and an output category. However, they require access to the internals of the base model, such as the gradients of the output with respect to the input, intermediate feature maps, or the network's weights. Many methods are also limited to certain network architectures and/or layer types~\cite{Zhou2016Learning}. In this paper, we advocate for a more general approach that can produce a saliency map for an arbitrary network without requiring access to its internals and does not require re-implementation for each network architecture. LIME~\cite{Ribeiro2016Should} offers such a black-box approach by drawing random samples around the instance to be explained and fitting an approximate linear decision model. However, its saliency is based on superpixels, which may not capture correct regions (see Fig.~\ref{fig:LIME}.) 

\begin{figure}[!t]
    \centering
    \vspace{-1mm}
    \includegraphics[width=0.9\linewidth]{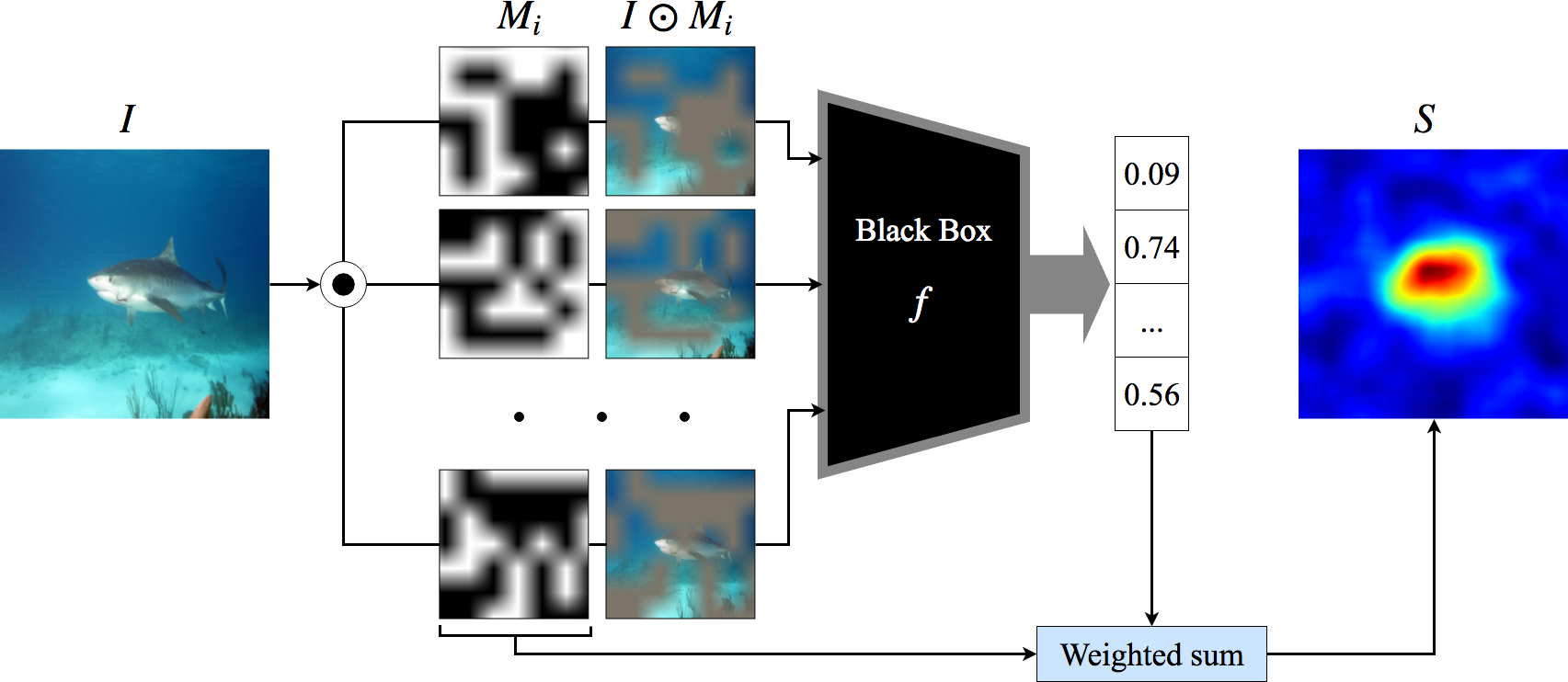}
    \caption{\small Overview of RISE: Input image $I$ is element-wise multiplied with random masks $M_i$ and the masked images are fed to the base model. The saliency map is a linear combination of the masks where the weights come from the score of the target class corresponding to the respective masked inputs.}
    \vspace{-3mm}
    \label{fig:Approach}
\end{figure}

We propose a new black-box approach for estimating pixel saliency called \textit{Randomized Input Sampling for Explanation (RISE)}. Our approach is general and applies to any off-the-shelf image network, treating it as a complete black box and not assuming access to its parameters, features or gradients.
The key idea is to probe the base model by sub-sampling the input image via random masks and recording its response to each of the masked images.
The final importance map is generated as a linear combination of the random binary masks where the combination weights come from the output probabilities predicted by the base model on the masked images (See Fig.~\ref{fig:Approach}).
This seemingly simple yet surprisingly powerful approach allows us to peek inside an arbitrary network without accessing any of its internal structure.
Thus, RISE is a true black-box explanation approach which is conceptually different from mainstream white-box saliency approaches such as GradCAM~\cite{Selvaraju2017Gradcam} and, in principle, is generalizable to base models of any architecture.
    
Another key contribution of our work is to propose causal metrics to evaluate the produced explanations.
Most explanation approaches are evaluated in a human-centered way, where the generated saliency map is compared to the ``ground truth''  regions or bounding boxes drawn by humans in localization datasets~\cite{Selvaraju2017Gradcam, zhang2017EB}.
Some approaches also measure human trust or reliability on the explanations~\cite{Ribeiro2016Should, Selvaraju2017Gradcam}.
Such evaluations not only require a lot of human effort but, importantly, are unfit for evaluating whether the explanation is the \textit{true cause} of the model's decision. They only capture how well the explanations imitate the human-annotated importance of the image regions.
But an AI system could behave differently from a human and learn to use cues from the background (\eg, using grass to detect cows) or other cues that are non-intuitive to humans. Thus, a human-dependent metric cannot evaluate the correctness of an explanation that aims to extract the underlying decision process from the network.   
    
Motivated by~\cite{Fong2017Interpretable}, we propose two automatic evaluation metrics: \textit{deletion} and \textit{insertion}. 
The deletion metric measures the drop in the probability of a class as important pixels (given by the saliency map) are gradually removed from the image.
A sharp drop, and thus a small area under the probability curve, are indicative of a good explanation.
Fig.~\ref{fig:ExplanationAndSelectivity} shows plots produced by different explanation techniques for an image containing `goldfish', where the total Area Under Curve (AUC) value is the smallest for our RISE model, indicating a more causal explanation.
The insertion metric, on the other hand, captures the importance of the pixels in terms of their ability to \textit{synthesize} an image and is measured by the rise in the probability of the class of interest as pixels are added according to the generated importance map.
We argue that these two metrics not only alleviate the need for large-scale human evaluation or annotation effort, but are also better at assessing causal explanations by being human agnostic.
For the sake of completeness, we also compare the performance of our method to state-of-the-art explanation models in terms of a human-centric evaluation metric.

\section{Related work}
\label{sec:related}
\vspace{-1mm}
The importance of producing explanations has been extensively studied in multiple fields, within and outside machine learning.
Historically, representing knowledge using rules or decision trees~\cite{Swartout1981Producing, Swartout1993Explanation} has been found to be interpretable by humans. Another line of research focused on approximating the less interpretable models (\eg, neural network, non-linear SVMs \etc) with simple, interpretable models such as decision rules or sparse linear models~\cite{Thrun1995Extracting, Craven1996Extracting}.
In a recent work Ribeiro \textit{et. al.}~\cite{Ribeiro2016Should}, fits a more interpretable approximate linear decision model (LIME) in the vicinity of a particular input.
Though the approximation is fairly good locally, for a sufficiently complex model, a linear approximation may not lead to a faithful representation of the non-linear model. The LIME model can be applied to black-box networks like our approach, but its reliance on superpixels leads to inferior importance maps as shown in our experiments.

To explain classification decisions in images, previous works either visually ground image regions that strongly support the decision~\cite{Selvaraju2017Gradcam, Park2018Multimodal} or generate a textual description of why the decision was made~\cite{Hendricks2016Generating}.
The visual grounding is generally expressed as a saliency or importance map which shows the importance of each pixel towards the model's decision.
Existing approaches to deep neural network explanation either design `interpretable' network architectures or attempt to explain or `justify' decisions made by an existing model.

Within the class of interpretable architectures, Xu \textit{et. al.}~\cite{Xu2015ShowAttention}, proposed an interpretable image captioning system by incorporating an attention network which learns where to look next in an image before producing each word of the caption. A neural module network is employed in~\cite{Andreas2016Learning, Hu2017Learning} to produce the answers to visual question-answering problems in an interpretable manner by learning to divide the problem into subproblems.
However, these approaches achieve interpretability by incorporating changes to a white-box base model and are constrained to use specific network architectures.

Neural justification approaches attempt to justify the decision of a base model.
\textit{Third-person} models~\cite{Hendricks2016Generating, Park2018Multimodal} train additional models from human annotated `ground truth' reasoning in the form of saliency maps or textual justifications. The success of such methods depends on the availability of tediously labeled ground-truth explanations, and they do not produce high-fidelity explanations.
Alternatively, \textit{first-person} models~\cite{Zhou2016Learning, Selvaraju2017Gradcam, Fong2017Interpretable, Cao2015Look} aim to generate explanations providing evidence for the model's underlying decision process without using an additional model. In our work, we focus on producing a first-person justification.

Several approaches generate importance maps by isolating contributions of image regions to the prediction.
In one of the early works~\cite{Zeiler2013Visualizing}, Zeiler \textit{et al.} visualize the internal representation learned by CNNs using deconvolutional networks.
Other approaches~\cite{Simonyan2013Deep, Yosinski2014Understanding, Nguyen2016Synthesizing} have tried to synthesize an input (an image) that highly activates a neuron.
The Class Activation Mapping (CAM) approach~\cite{Zhou2016Learning} achieves class-specific importance of each location of an image by computing a weighted sum of the feature activation values at that location across all channels.
However, the approach can only be applied to a particular kind of CNN architecture where a global average pooling is performed over convolutional feature map channels immediately prior to the classification layer.
Grad-CAM~\cite{Selvaraju2017Gradcam} extends CAM by weighing the feature activation values at every location with the average gradient of the class score (w.r.t. the feature activation values) for every feature map channel.
Zhang \textit{et al.}~\cite{zhang2017EB} introduce a probabilistic winner-take-all strategy to compute top-down importance of neurons towards model predictions.
Fong \textit{et al.}~\cite{Fong2017Interpretable} and Cao \textit{et al.}~\cite{Cao2015Look} learn a perturbation mask that maximally affects the model's output by backpropagating the error signals through the model.
However, all of the above methods~\cite{Simonyan2013Deep, Yosinski2014Understanding, Nguyen2016Synthesizing, Zhou2016Learning, Selvaraju2017Gradcam, zhang2017EB,Fong2017Interpretable, Cao2015Look} assume access to the internals of the base model to obtain feature activation values, gradients or weights.
RISE is a more general framework as the importance map is obtained with access to only the input and output of the base model.

\section{Randomized Input Sampling for Explanation (RISE)}
\label{sec:approach}
\vspace{-1mm}
One way to measure the importance of an image region is to obscure or `perturb' it and observe how much this affects the black box decision. For example, this can be done by setting pixel intensities to zero~\cite{Zeiler2013Visualizing, Fong2017Interpretable,Ribeiro2016Should}, blurring the region~\cite{Fong2017Interpretable} or by adding noise.
In this work we estimate the importance of pixels by dimming them in random combinations, reducing their intensities down to zero. We model this by multiplying an image with a $[0,1]$ valued mask. 
The mask generation process is described in detail in section~\ref{subsec:mask}.
    
    \subsection{Random Masking}
    
    Let $f:\mathcal{I}\rightarrow\mathbb{R}$ be a black-box model, that for a given input from $\mathcal{I}$ produces scalar confidence score. 
    In our case, $\mathcal{I}$ is the space of color images $\mathcal{I}=\{I\mid I:\Lambda\rightarrow\mathbb{R}^3\}$ of size $H\times W$ ($\Lambda=\{1,\ldots, H\}\times\{1,\ldots,W\}$), where every image $I$ is a mapping from coordinates to three color values. For example, $f$ may be a classifier that produces the probability that object of some class is present in the image, or a captioning model that outputs the probability of the next word given a partial sentence. 

Let $M:\Lambda \rightarrow \{0,1\}$ be a random binary mask with distribution $\mathcal{D}$. Consider the random variable $f(I\odot M)$, where $\odot$ denotes element-wise multiplication. First, the image is masked by preserving only a subset of pixels. Then, the confidence score for the masked image is computed by the black box. We define importance of pixel $\lambda\in\Lambda$ as the expected score over all possible masks $M$ conditioned on the event that pixel $\lambda$ is observed, \ie, $M(\lambda)=1$:
\begin{equation}
\label{eq:importance}
S_{I,f}(\lambda) = \mathbb{E}_M\left[f(I\odot M)\mid M(\lambda)=1\right].
\end{equation}
The intuition behind this is that $f(I\odot M)$ is high when pixels preserved by mask $M$ are important. 

Eq.~(\ref{eq:importance}) can be rewritten as a summation over mask $m:\Lambda\rightarrow\{0,1\}$:
\begin{align}
    \label{eq:inter}
    S_{I,f}(\lambda) &= \sum_{m}f(I\odot m)P[M=m \mid M(\lambda)=1]\\
	&= \frac{1}{P[M(\lambda)=1]}\sum_{m} f\left(I\odot m\right)P[M=m,\ M(\lambda)=1]. \nonumber
\end{align}
\indent Now, 
\begin{align}
    \label{eq:joint}
    P[M=m,\ M(\lambda)=1] &= 
    \begin{cases}
        0, &\text{ if }m(\lambda)=0,\\
        P[M=m], &\text{ if }m(\lambda)=1.
    \end{cases}\\
    &= m(\lambda)P[M=m].\nonumber
\end{align}

Substituting $P[M=m,\ M(\lambda)=1]$ from (\ref{eq:joint}) in (\ref{eq:inter}),
\begin{align}
    S_{I,f}(\lambda) &= \frac{1}{P[M(\lambda)=1]}\sum_{m} f\left(I\odot m\right) \cdot m(\lambda)\cdot P[M=m] \label{eq:final_scalar}.
\end{align}

It can be written in matrix notation, combined with the fact that $P[M(\lambda)=1]=\mathbb{E}[M(\lambda)]$:
\begin{align}
    S_{I,f} &= \frac{1}{\mathbb{E}[M]}\sum_{m} f\left(I\odot m\right) \cdot m\cdot P[M=m] \label{eq:final_matrix}.
\end{align}

Thus, saliency map can be computed as a weighted sum of random masks, where weights are the probability scores, that masks produce, adjusted for the distribution of the random masks.

We propose to generate importance maps by empirically estimating the sum in equation~(\ref{eq:final_matrix}) using Monte Carlo sampling. To produce an importance map, explaining the decision of model $f$ on image $I$, we sample set of masks $\{M_1,\ldots, M_N\}$ according to $\mathcal{D}$ and probe the model by running it on masked images $I\odot M_i$, $i=1,\ldots, N$.
Then, we take the weighted average of the masks where the weights are the confidence scores $f(I\odot M_i)$ and normalize it by the expectation of $M$:
\begin{align}
\label{main}
S_{I, f}(\lambda) &\overset{\mathrm{MC}}{\approx} \frac 1{\mathbb{E}[M]\cdot N}\sum_{i=1}^Nf(I\odot M_i)\cdot M_i(\lambda).
\end{align}

Note that our method does not use any information from inside the model and thus, is suitable for explaining black-box models. 

\subsection{Mask generation}
\vspace{-1mm}
\label{subsec:mask}

Masking pixels independently may cause adversarial effects: a slight change in pixel values may cause significant variation in the model's confidence scores. Moreover, generating masks by independently setting their elements to zeros and ones will result in mask space of size $2^{H\times W}$. A larger space size requires more samples for a good estimation in equation (\ref{main}).
    
To address these issues we first sample smaller binary masks and then upsample them to larger resolution using bilinear interpolation. Bilinear upsampling does not introduce sharp edges in $I\odot M_i$ as well as results in a smooth importance map $S$. After interpolation, masks $M_i$ are no longer binary, but have values in $[0,1]$. Finally, to allow more flexible masking, we shift all masks by a random number of pixels in both spatial directions.

Formally, mask generation can be summarized as:
\begin{enumerate}[nosep]
	\item Sample $N$ binary masks of size $h\times w$ (smaller than image size $H\times W$) by setting each element independently to $1$ with probability $p$ and to $0$ with the remaining probability.
	\item Upsample all masks to size $(h+1)C_H\times(w+1)C_W$ using bilinear interpolation, where $C_H\times C_W=\lfloor H/h\rfloor\times\lfloor W/w\rfloor$ is the size of the cell in the upsampled mask.
	\item Crop areas $H\times W$ with uniformly random indents from $(0,0)$ up to $(C_H, C_W)$.
\end{enumerate}
\vspace{-2mm}

\section{Experiments}
\vspace{-1mm}
\uline{Datasets and Base Models}:
We evaluated RISE on 3 publicly available object classification datasets, namely, PASCAL VOC07~\cite{Everingham2010Pascal}, MSCOCO2014~\cite{Lin2014Microsoft} and ImageNet~\cite{Russakovsky2015Imagenet}.
Given a base model, we test importance maps generated by different explanation methods for a target object category present in images from the VOC and MSCOCO datasets.
For the ImageNet dataset, we test the explanation generated for the top probable class of the image.
We chose the particular versions of the VOC and MSCOCO datasets to compare fairly with the state-of-the-art reporting on the same datasets and same base models.
For these two datasets, we used ResNet50~\cite{He2016Deep} and VGG16~\cite{Simonyan2013Very} networks trained by~\cite{zhang2017EB} as base models.
For ImageNet, the same base models were downloaded from the PyTorch model zoo~\footnote{\url{https://github.com/pytorch/vision}}.
    
\subsection{Evaluation Metrics}
\label{subsec:EvalMetric}
Despite a growing body of research focusing on explainable machine learning, there is still no consensus about how to measure the explainability of a machine learning model~\cite{Poursabzi2018Manipulating}.
As a result, human evaluation has been the predominant way to assess model explanation by measuring it from the perspective of transparency, user trust or human comprehension of the decisions made by the model~\cite{Herman2018Promise}.
Existing justification methods~\cite{zhang2017EB, Selvaraju2017Gradcam} have evaluated saliency maps by their ability to localize objects.
However, localization is merely a proxy for human explanation and may not correctly capture what \textit{causes} the base model to make a decision irrespective of whether the decision is right or wrong as far as the proxy task is concerned.
As a typical example, let us consider an image of a car driving on a road.
Evaluating an explanation against the localization bounding box of the car does not give credit for (in fact discredits) correctly capturing `road' as a possible cause behind the base model's decision of classifying the image as that of a car.
We argue that keeping humans out of the loop for evaluation makes it more fair and true to the classifier's own \emph{view} on the problem rather than representing a human's view.
Such a metric is not only objective (free from human bias) in nature but also saves time and resources.

\begin{figure}[!t]
    \centering
    \vspace{-1mm}
    \includegraphics[width=0.8\linewidth]{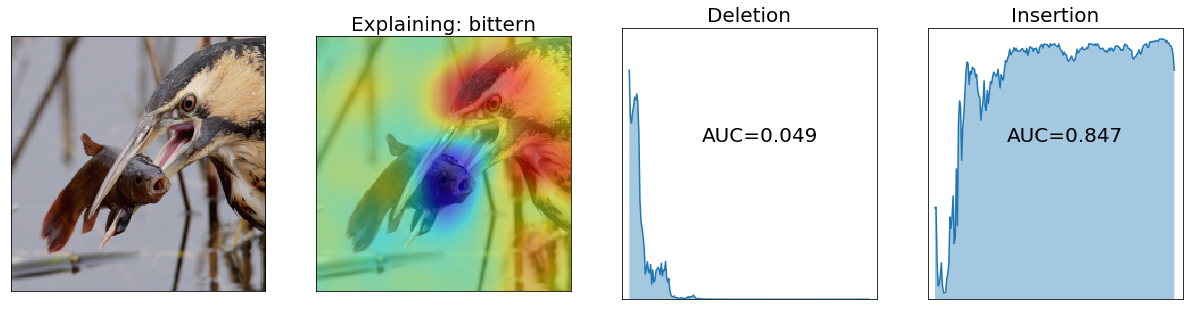}\\
    \includegraphics[width=0.8\linewidth]{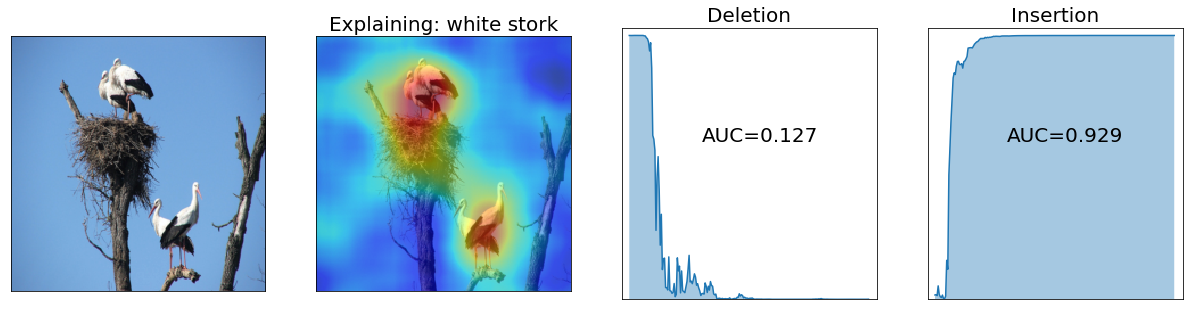}\\
    \caption{\small RISE-generated importance maps (second column) for two representative images (first column) with deletion (third column) and insertion  (fourth column) curves.}
    \vspace{-3mm}
    \label{fig:additional_examples}
\end{figure}

\uline{Causal metrics for explanations:}
\label{subsubsec:causal}
To address these issues, we propose two automatic evaluation metrics: \textit{deletion} and \textit{insertion}, motivated by~\cite{Fong2017Interpretable}.
The intuition behind the \textit{deletion metric} is that the removal of the `cause' will force the base model to change its decision.
Specifically, this metric measures a decrease in the probability of the predicted class as more and more important pixels are removed, where the importance is obtained from the importance map.
A sharp drop and thus a low area under the probability curve (as a function of the fraction of removed pixels) means a good explanation.
The \textit{insertion metric}, on the other hand, takes a complementary approach.
It measures the increase in probability as more and more pixels are introduced, with higher AUC indicative of a better explanation.

There are several ways of removing pixels from an image~\cite{Dabkowski2017Real}, \eg, setting the pixel values to zero or any other constant gray value, blurring the pixels or even cropping out a tight bounding box.
The same is true when pixels are introduced, \textit{e.g.}, they can be introduced to a constant canvas or by starting with a highly blurred image and gradually unblurring regions.
All of these approaches have different pros and cons. A common issue is the introduction of spurious evidence which can fool the classifier.
For example, if pixels are introduced to a constant canvas and if the introduced region happens to be oval in shape, the classifier may classify the image as a `balloon' (possibly a printed balloon) with high probability.
This issue is less severe if pixels are introduced to an initially blurred canvas as blurring takes away most of the finer details of an image without exposing it to sharp edges as image regions are introduced.
This strategy gives higher scores for all methods, so we adopt it for insertion.
For deletion, the aim is to fool the classifier as quickly as possible and blurring small regions instead of setting them to a constant gray level does not help.
This is because a good classifier is usually able to fill in the missing details quite remarkably from the surrounding regions and from the small amount of low-frequency information left after blurring a tiny region.
As a result, we set the image regions to constant values when removing them for the deletion metric evaluation.
We used the same strategies for all the existing approaches with which we compared our method in terms of these two metrics.

\uline{Pointing game:}
\label{subsubsec:Pointing}
We also evaluate explanations in terms of a human evaluation metric, the \textit{pointing game} introduced in~\cite{zhang2017EB}.
If the highest saliency point lies inside the human-annotated bounding box of an object, it is counted as a hit.
The pointing game accuracy is given by $\frac{\# Hits}{\# Hits + \# Misses}$, averaged over all target categories in the dataset. For a classification model that learns to rely on objects, this metric should be high for a good explanation.

\vspace{-2mm}
\subsection{Experimental Results}
\label{subsec:experiments}

\noindent\uline{Experimental Settings}:
The binary random masks are generated with equal probabilities for 0's and 1's.
For different CNN classifiers, we empirically select different numbers of masks, in particular, we used 4000 masks for the VGG16 network and 8000 for ResNet50.
We have used $h=w=7$ and $H=W=224$ throughout.
All the results used for comparison were either taken from published works or by running the publicly available code on datasets for which reported results could not be obtained.

\begin{table}
	\centering
	\caption{\small Comparative evaluation in terms of deletion (lower is better) and insertion (higher is better) scores on ImageNet dataset. Except for Grad-CAM, the rest are black-box explanation models.}
	\label{table:del-ins}
	\footnotesize
	\begin{tabular}{l|c|c||c|c}
		\hline
		\multirow{2}{*}{Method} & \multicolumn{2}{c}{ResNet50} & \multicolumn{2}{c}{VGG16}\\
		\cline{2-5}
		& Deletion & Insertion & Deletion & Insertion \\
		\hline
		Grad-CAM~\cite{Selvaraju2017Gradcam} & $0.1232$ & $0.6766$ & $0.1087$ & $0.6149$\\
		\hline
		Sliding window~\cite{Zeiler2013Visualizing} & $0.1421$ & $0.6618$ & $0.1158$ & $0.5917$ \\
		LIME~\cite{Ribeiro2016Should}        & $0.1217$ & $0.6940$ & $0.1014$ & $0.6167$ \\
		RISE (ours) & $\mathbf{0.1076\pm 0.0005}$ & $\mathbf{0.7267 \pm 0.0006}$ & $\mathbf{0.0980\pm 0.0025}$ & $\mathbf{0.6663\pm 0.0014}$ \\ 
		\hline
	\end{tabular}
\vspace{-5mm}
\end{table}

\noindent\uline{Deletion and Insertion scores}:
Table \ref{table:del-ins} shows a comparative evaluation of RISE with other state-of-the-art approaches in terms of both deletion and insertion metrics on val split of ImageNet.
RISE reports an average value with associated standard deviations for 3 independent runs.
The sliding window approach~\cite{Zeiler2013Visualizing} systematically occludes fixed size image regions and probes the model with the perturbed image to measure the importance of the occluded region.
We used a sliding window of size $64\times64$ with stride $8$.
For LIME~\cite{Ribeiro2016Should}, the number of samples was set to $1000$ (taken from the code).
For this experiment, we used the ImageNet classification dataset where no ground truth segmentation or localization mask is provided and thus explainability performance can only be measured via automatic metrics like deletion and insertion.
For both the base models and according to both the metrics, RISE provides better performance, outperforming even the white-box Grad-CAM method.
The values are better for ResNet50 which is intuitive as it is a better classification model than VGG16.
However, due to increased number of forward passes, RISE is heavy in computation. This can potentially be addressed by intelligently sampling fewer number of random masks which is kept as a future work. RISE, sometimes, provides noisy importance maps due to sampling approximation especially in presence of objects with varying sizes.
Fig.~\ref{fig:additional_examples} shows examples of RISE-generated importance maps along with the deletion and insertion curves.
The appendix contains additional visual examples including a few noisy importance maps.

\noindent\uline{Pointing game accuracy}:
The performance in terms of pointing game accuracy is shown in Table~\ref{tab:PointingGameEval} for the test split of PASCAL VOC07 and val split of MSCOCO2014 datasets. In this table, RISE is the only black-box method.
The base models are obtained from~\cite{zhang2017EB} and thus we list the pointing game accuracies reported in the paper.
RISE reports an average value of 3 independent runs; low standard deviation values indicate the robustness of the proposed approach against the randomness of the masks.
For VGG16, RISE performs consistently better than all of the white-box methods with a significantly improved performance for the VOC dataset.
For the deeper ResNet50 network with residual connections, RISE does not have the highest pointing accuracy but comes close.
We stress again that good pointing accuracy may not correlate with actual causal processes in a network, however, RISE is competitive despite being black-box and more general than methods like CAM, which is only applicable to architectures without fully-connected layers.

\vspace{-3mm}
\subsection{RISE for Captioning}
\label{sec:captioning}

RISE can easily be extended to explain captions for any image description system.
Some existing works use a separate attention network~\cite{Xu2015ShowAttention} or assume access to feature activations~\cite{zhang2017EB} and/or gradient values~\cite{Selvaraju2017Gradcam} to ground words in an image caption.
The most similar to our work is Ramanishka \etal~\cite{Ramanishka2017Top} where the base model is probed with conv features from small patches of the input image to estimate its importance for each word in the caption.
However, our approach is not constrained to a single fixed size patch and is thus less sensitive to object sizes as well as better at capturing additional context that may be present in the image.
We provide a small example of RISE being applied for explaining image caption.
    
\begin{figure}[!t]
    \centering
    \subfigure["A horse and carriage on a city street."]{
    	\label{subfig:Original}
    	\includegraphics[width=0.23\linewidth]{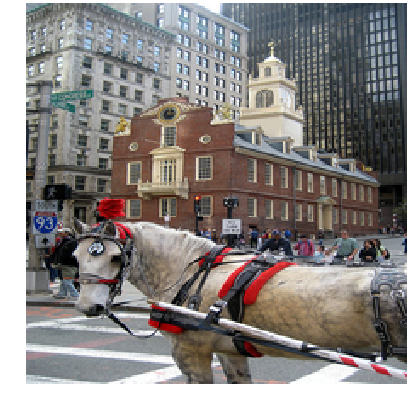}}
    \subfigure["A \textcolor{red}{horse}..."]{
    	\label{subfig:Horse}
    	\includegraphics[width=0.23\linewidth]{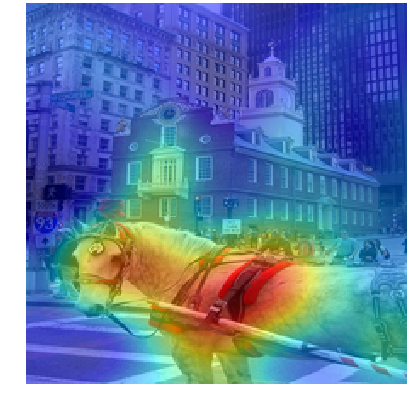}}
    \subfigure["A horse and \textcolor{red}{carriage}..."]{
    	\label{subfig:Carriage}
    	\includegraphics[width=0.23\linewidth]{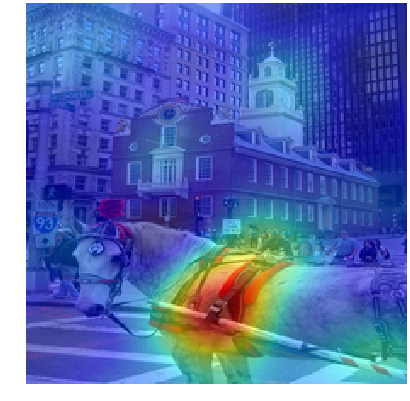}}
    \subfigure["\textcolor{red}{White}..."]{
    	\label{subfig:White}
    	\includegraphics[width=0.23\linewidth]{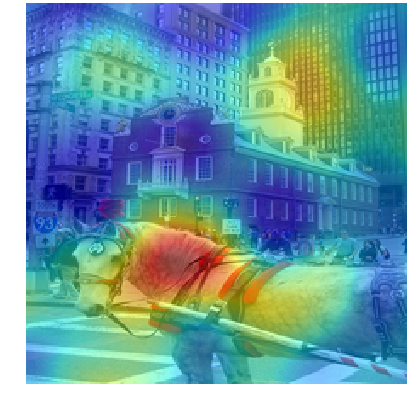}}
    \caption{\small Explanations of image captioning models. \protect\subref{subfig:Original} is the image with the caption generated by~\cite{Donahue2015Long}. \protect\subref{subfig:Horse} and \protect\subref{subfig:Carriage} show the importance map generated by RISE for two words `horse' and `carriage' respectively from the generated caption. \protect\subref{subfig:White} shows the importance map for an arbitrary word `white'.}
    \label{fig:Captions}
\end{figure}

We take a base captioning model~\cite{Donahue2015Long} that models the probability of the next word $w_{k}$ given a partial sentence $s=(w_1, \ldots, w_{k-1})$ and an input image $I$:
\begin{equation}
    \vspace{-1mm}
    f(I, s, w_k) = P[w_k|I, w_1, \ldots, w_{k-1}]
    \vspace{-1mm}
\end{equation}
We probe the base model by running it on a set of $N$ randomly masked inputs $f(I\odot M_i, s, w_k)$ and computing saliency as $\frac{1}{N\cdot\mathbb{E}[M] }\sum_{i=1}^Nf(I\odot M_i, s, w_k)\cdot M_i$ for each word in $s$.
Input sentence $s$ can be any arbitrary sentence including the caption generated by the base model itself.
Three such explanation instances for MSCOCO image are shown in Fig.~\ref{fig:Captions}.
    
\begin{table*}
\centering
\caption{\small Mean accuracy (\%) in the pointing game. Except for RISE, the rest require white-box model.}
\label{tab:PointingGameEval}
\begin{adjustwidth}{-1mm}{-2mm}
\footnotesize
\begin{tabulary}{1\linewidth}{p{12mm}|p{1cm}||p{8mm}|p{12mm}|C|C|C|C}
    \hline
	\hspace{-2mm}Base model & Dataset & AM~\cite{Simonyan2013Deep} & Deconv~\cite{Zeiler2013Visualizing} & CAM~\cite{Zhou2016Learning} & MWP~\cite{zhang2017EB} & c-MWP~\cite{zhang2017EB} & RISE \\
    \hhline{=|=#=|=|=|=|=|=}
    
    \hspace{-2mm}\multirow{2}{*}{VGG16} & VOC 		& 76.00 & 75.50 & - & 76.90 & 80.00 & $\mathbf{87.33}\pm 0.49$ \\
    					   				& MSCOCO 	& 37.10 & 38.60 & - & 39.50 & 49.60 & $\mathbf{50.71}\pm 0.10$ \\
	\hline
    \hspace{-2mm}\multirow{2}{*}{Resnet50} & VOC 	& 65.80 & 73.00 & \textbf{90.60} & 80.90 & 89.20 & $88.94\pm 0.61$ \\
    & MSCOCO 	& 30.40 & 38.2 & \textbf{58.4} & 46.8 & 57.4 & $55.58\pm 0.51$ \\
	\hline
\end{tabulary}
\end{adjustwidth}
\end{table*}

\section{Conclusion}
\label{sec:conclusion}
This paper presented RISE, an approach for explaining black-box models by estimating the importance of input image regions for the model's prediction. Despite its simplicity and generality, the method outperforms existing explanation approaches in terms of automatic causal metrics and performs competitively in terms of the human-centric pointing metric.
Future work will be to exploit the generality of the approach for explaining decisions made by complex networks in video and other domains.

\noindent{\bf Acknowledgement}: This work was partially supported by the DARPA XAI program.

\bibliography{XAI}

\newpage
\begin{appendices}
\section{Algorithms to compute causal metrics}

Algorithm to compute deletion score.
\begin{algorithm}
\small
\caption{}\label{alg:deletion}
\begin{algorithmic}[1]
\Procedure{Deletion}{}
\State \textbf{Input:} black box $f$, image $I$, importance map $S$, number of pixels $N$ removed per step
\State \textbf{Output:} deletion score $d$
\State $n\gets 0$
\State $h_n\gets f(I)$
\While {$I$ has non-zero pixels}
\State According to $S$, set next $N$ pixels in $I$ to 0
\State $n \gets n+1$
\State $h_n\gets f(I)$
\EndWhile
\State $d \gets$\texttt{AreaUnderCurve}$(h_i\ \textit{vs.}\ i/n,\quad \forall i=0,\ldots n)$
\State \Return $d$
\EndProcedure
\end{algorithmic}
\end{algorithm}

\noindent Algorithm to compute insertion score.
\begin{algorithm}
\small
\caption{}\label{alg:insertion}
\begin{algorithmic}[1]
\Procedure{Insertion}{}
\State \textbf{Input:} black box $f$, image $I$, importance map $S$, number of pixels $N$ removed per step
\State \textbf{Output:} insertion score $d$
\State $n\gets 0$
\State $I^\prime \gets$\texttt{Blur}$(I)$
\State $h_n\gets f(I)$
\While {$I\neq I^\prime$}
\State According to $S$, set next $N$ pixels in $I^\prime$ to corresponding pixels in $I$
\State $n \gets n+1$
\State $h_n\gets f(I^\prime)$
\EndWhile
\State $d \gets$\texttt{AreaUnderCurve}$(h_i\ \textit{vs.}\ i/n,\quad \forall i=0,\ldots n)$
\State \Return $d$
\EndProcedure
\end{algorithmic}
\end{algorithm}

\newpage

\section{More saliency maps and their scores}
\begin{figure}[!h]
	\centering
    \vspace{-1mm}
    \includegraphics[width=0.85\linewidth]{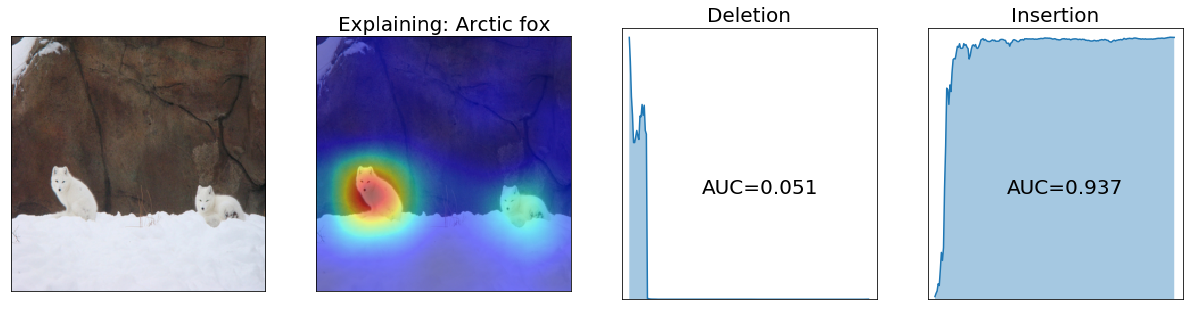}\\
    \includegraphics[width=0.85\linewidth]{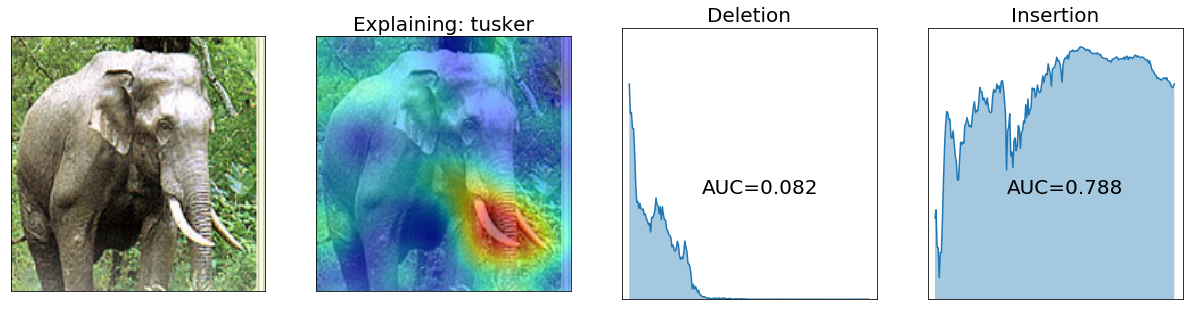}\\
    \includegraphics[width=0.85\linewidth]{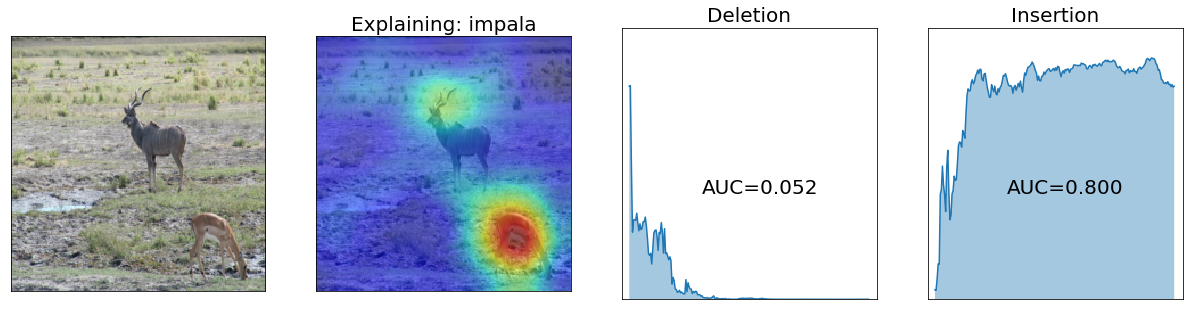}\\
    \includegraphics[width=0.85\linewidth]{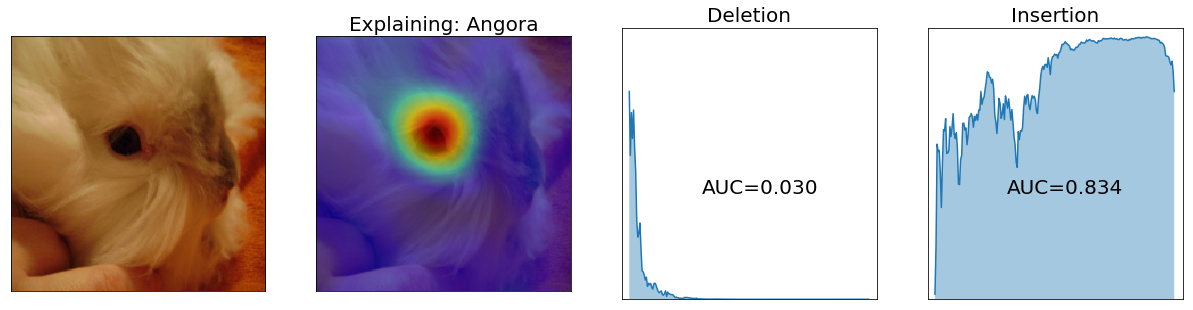}\\
    \includegraphics[width=0.85\linewidth]{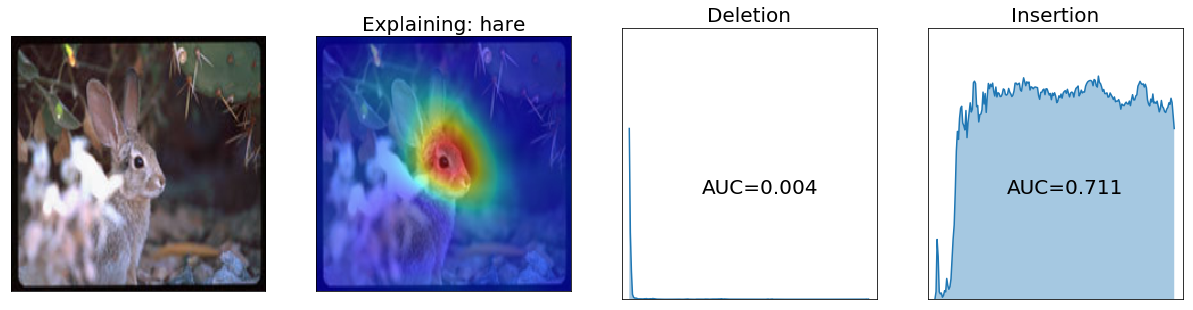}\\
    \includegraphics[width=0.85\linewidth]{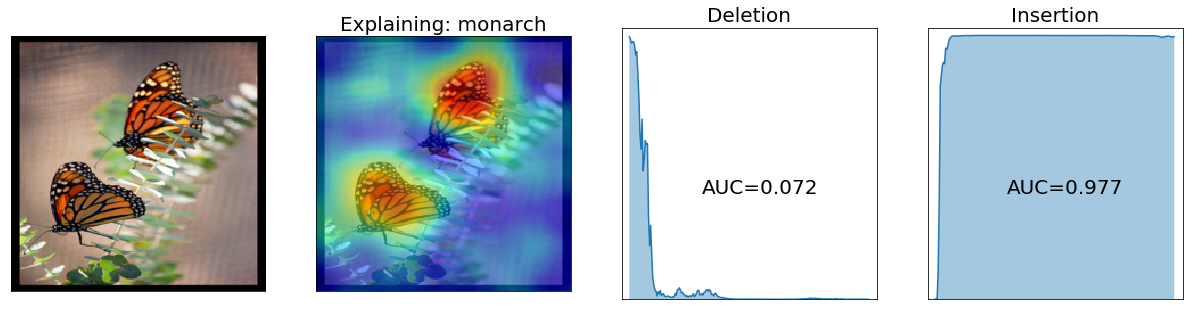}\\
    \caption{\small RISE-generated importance maps (second column) for representative images (first column) with deletion (third column) and insertion (fourth column) curves.}
    \vspace{-3mm}
\end{figure}

\newpage
\indent\begin{figure}[!h]
	\centering
    \vspace{-1mm}
    \includegraphics[width=0.85\linewidth]{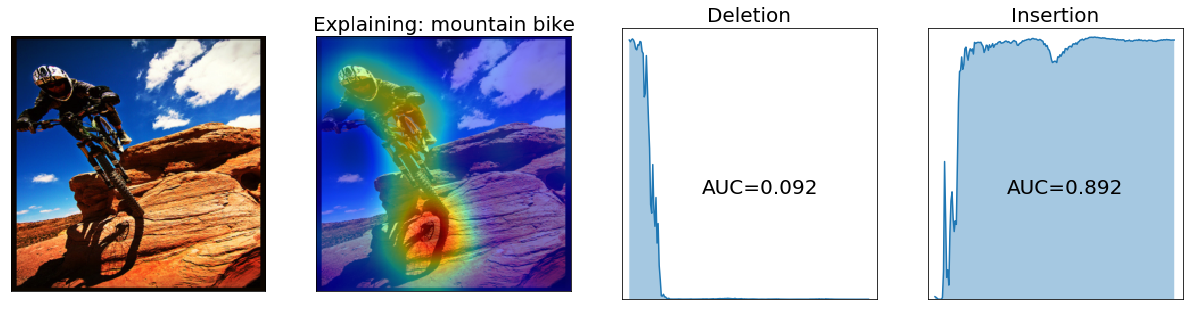}\\
    \includegraphics[width=0.85\linewidth]{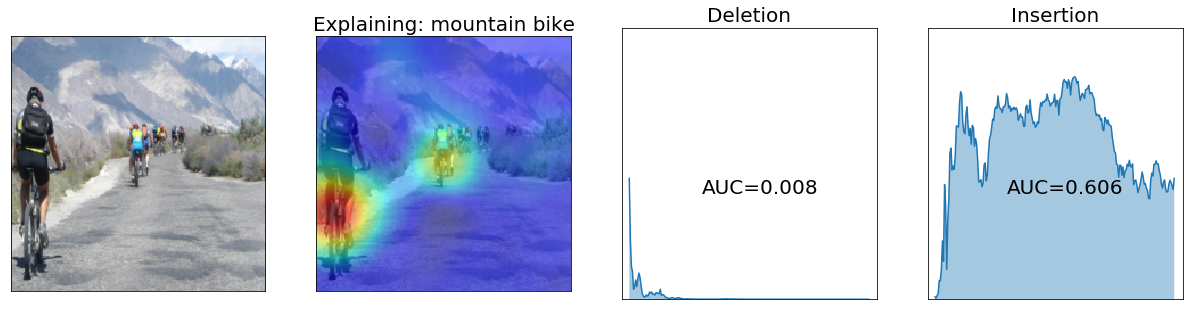}\\
    \includegraphics[width=0.85\linewidth]{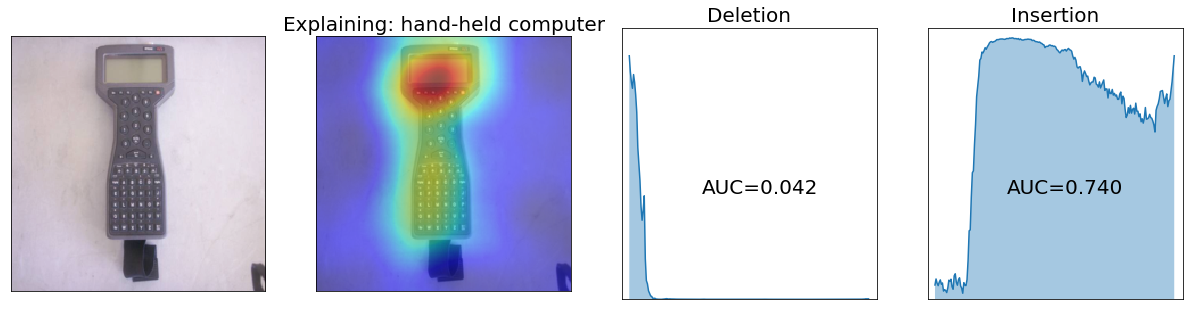}\\
    \includegraphics[width=0.85\linewidth]{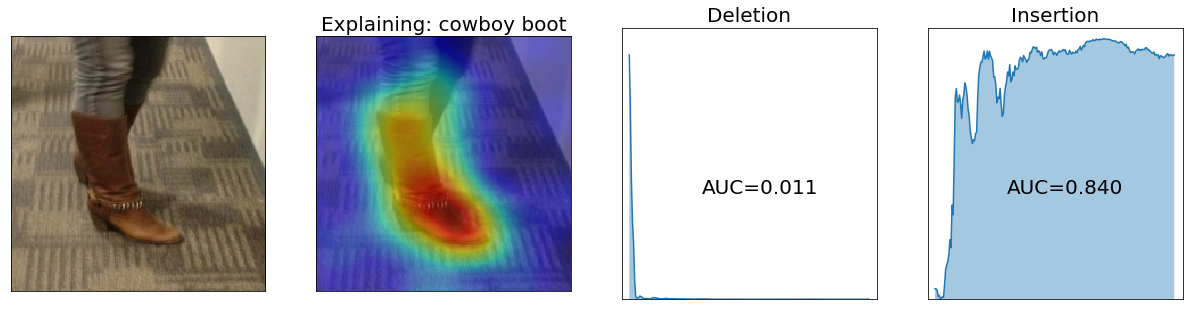}\\
    \includegraphics[width=0.85\linewidth]{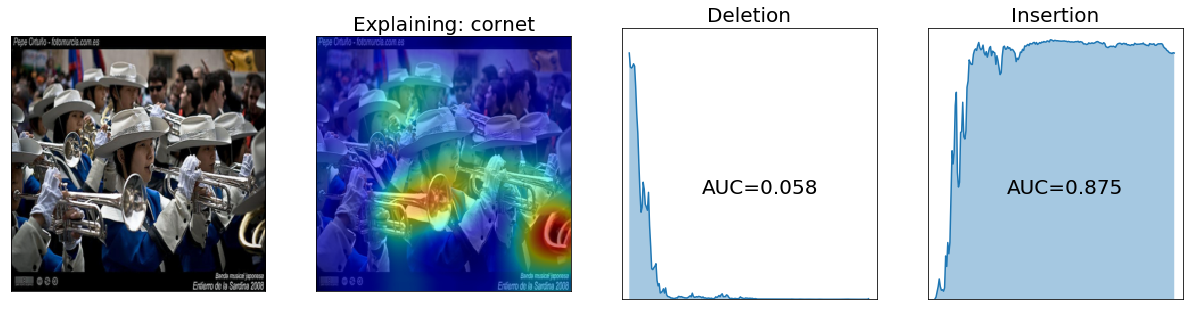}\\
    \includegraphics[width=0.85\linewidth]{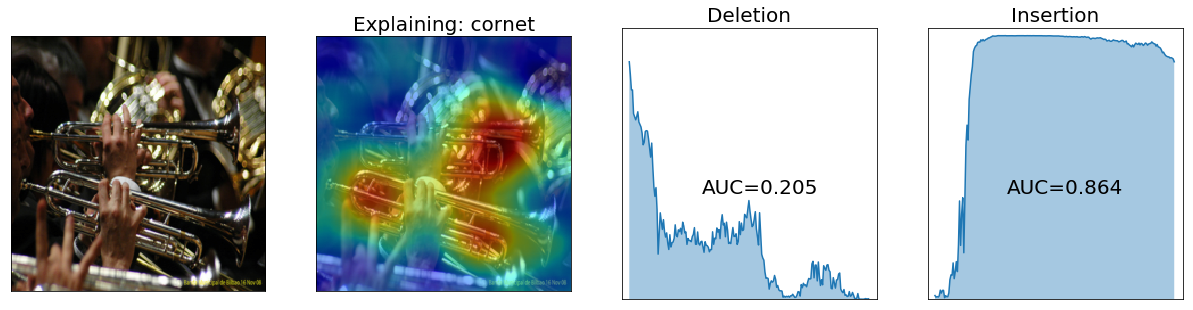}\\
    \caption{\small RISE-generated importance maps (second column) for representative images (first column) with deletion (third column) and insertion (fourth column) curves.}
    \vspace{-3mm}
\end{figure}

\newpage
\begin{figure}[!h]
	\centering
    \vspace{-1mm}
    \includegraphics[width=0.85\linewidth]{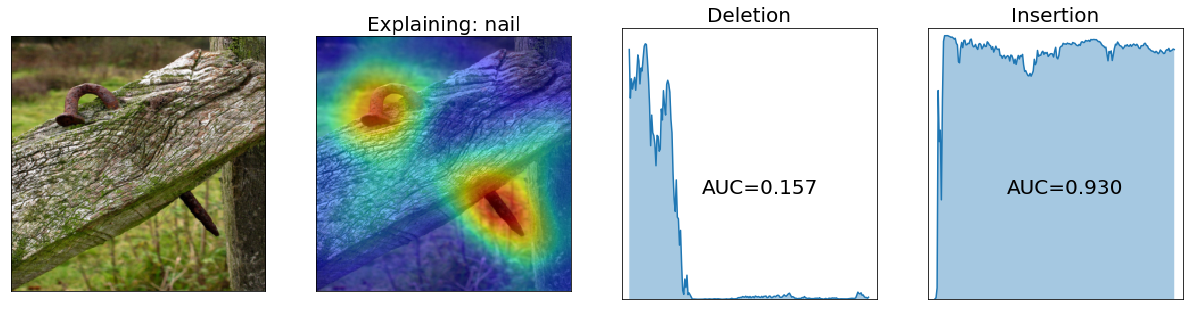}\\
    \includegraphics[width=0.85\linewidth]{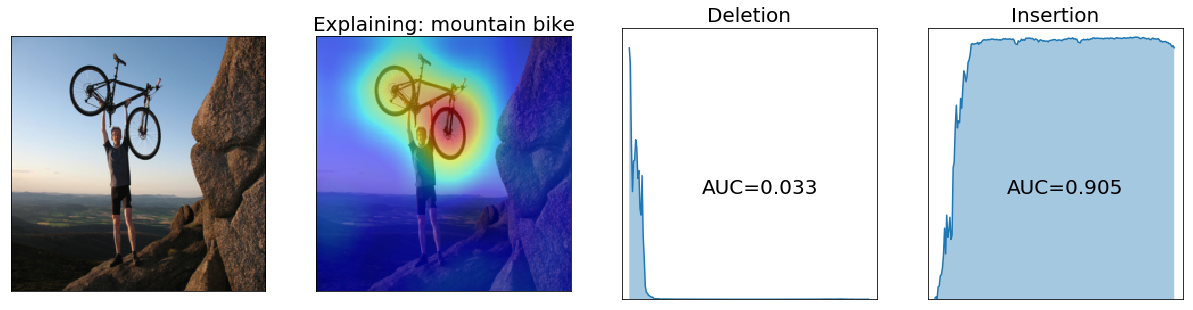}\\
    \includegraphics[width=0.85\linewidth]{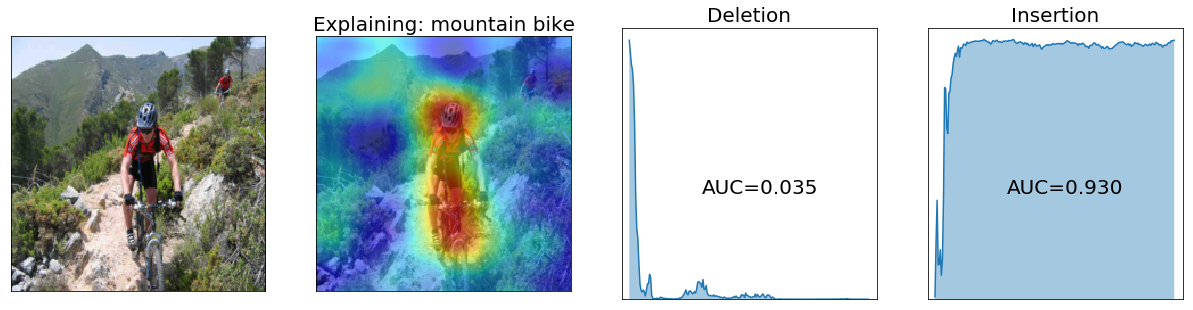}\\
    \caption{\small RISE-generated importance maps (second column) for representative images (first column) with deletion (third column) and insertion (fourth column) curves.}
    \vspace{-3mm}
\end{figure}

\begin{figure}[!h]
	\centering
    \vspace{-1mm}
    \includegraphics[width=0.85\linewidth]{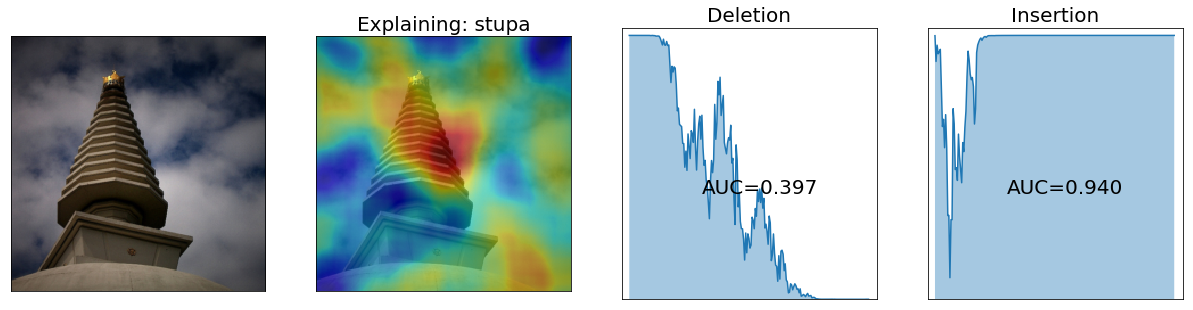}\\
    \includegraphics[width=0.85\linewidth]{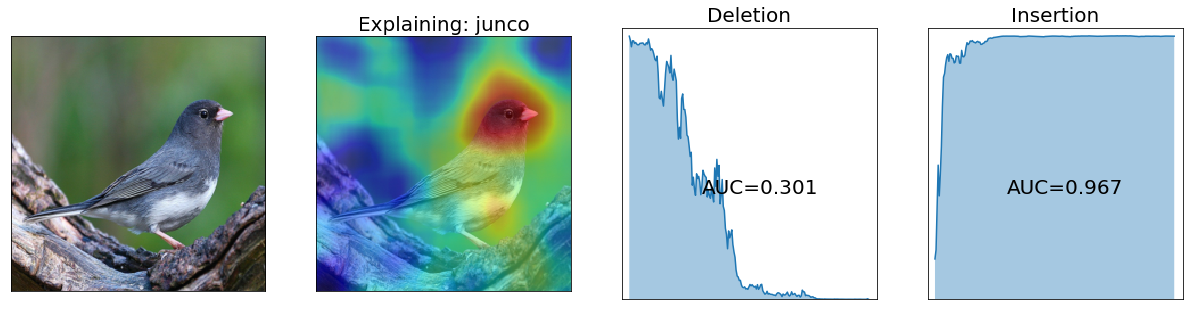}\\
    \includegraphics[width=0.85\linewidth]{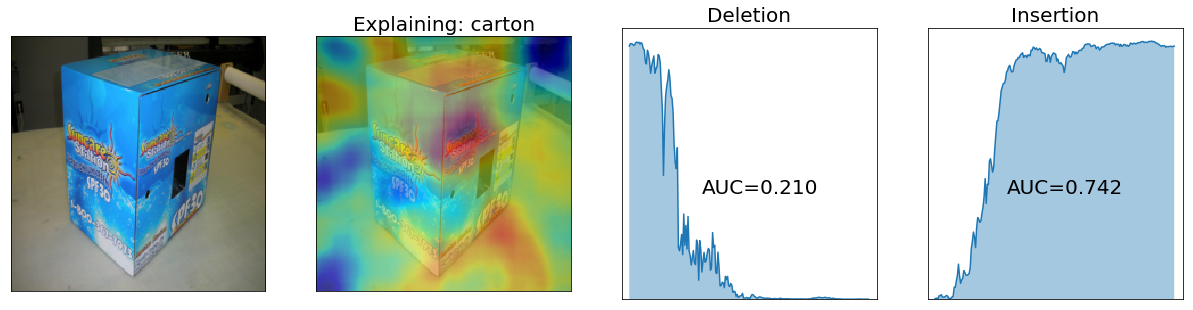}\\
    \caption{\small Failure cases. In some cases RISE does pick up more important features, but cannot get rid of the background noise (in part due to MC approximation with only a subset) like in rows 1 and 2.}
    \vspace{-3mm}
\end{figure}

\end{appendices}

\end{document}